\documentclass[letterpaper, 10 pt, conference]{ieeeconf}  

\IEEEoverridecommandlockouts                              

\usepackage{adjustbox}
\usepackage[font=small,belowskip=-1.5pt]{caption} 
\usepackage{tikz}
\usepackage[colorlinks=true]{hyperref}
\let\labelindent\relax
\usepackage[inline]{enumitem}

\usepackage{amssymb}
\usepackage{mathtools} 
\usepackage{bbm}

\usepackage{pifont}
\usepackage{alphalph}
\usepackage{amsmath}
\usepackage{booktabs}
\usepackage{multirow}
\usepackage{interval}
\usepackage{caption}
\usepackage{subcaption}

\definecolor{azure}{rgb}{0.0, 0.5, 1.0}

\usepackage{gensymb}

\newcommand{\ours}{\textsc{AGL-Net}}

\title{\LARGE \bf
\ours{}: Aerial-Ground Cross-Modal Global Localization with Varying Scales\\
}

\author{
Tianrui Guan$^{*}$\thanks{* Equal contribution.}
, Ruiqi Xian$^{*}$, Xijun Wang, Xiyang Wu, Mohamed Elnoor, Daeun Song, Dinesh Manocha
}
\begin{document}

\bstctlcite{IEEEexample:BSTcontrol}
\maketitle
\thispagestyle{empty}
\pagestyle{empty}

\begin{abstract}
We present \ours{}, a novel learning-based method for global localization using LiDAR point clouds and satellite maps.  \ours{} tackles two critical challenges: bridging the representation gap between image and points modalities for robust feature matching, and handling inherent scale discrepancies between global view and local view. To address these challenges, \ours{} leverages a unified network architecture with a novel two-stage matching design. The first stage extracts informative neural features directly from raw sensor data and performs initial feature matching.  The second stage refines this matching process by extracting informative skeleton features and incorporating a novel scale alignment step to rectify scale variations between LiDAR and map data. Furthermore, a novel scale and skeleton loss function guides the network toward learning scale-invariant feature representations, eliminating the need for pre-processing satellite maps. This significantly improves real-world applicability in scenarios with unknown map scales. To facilitate rigorous performance evaluation, we introduce a meticulously designed dataset within the CARLA simulator specifically tailored for metric localization training and assessment.
The code and data can be accessed at \href{https://github.com/rayguan97/AGL-Net}{https://github.com/rayguan97/AGL-Net}.


\end{abstract}


\section{Introduction}
Robotic navigation has been studied extensively in autonomous driving, mobile robotics and related applications. This includes new algorithms for many underlying problems, including collision avoidance~\cite{Sathyamoorthy2020DenseCAvoidRN, Sathyamoorthy2020FrozoneFP}, trajectory smoothness~\cite{10342393}, energy efficiency~\cite{kyaw2022energy}, off-road safety~\cite{ganav, 10161251, 9981942}, etc. Many approaches have been proposed for local and global planning. The latter include techniques based on using global waypoints and localization~\cite{beinhofer2013effective}. However, it is hard to develop global navigation methods over larger areas in real-world scenes. This is due to missing details in the representations used for the scene~\cite{guan2023crossloc3d, batalin2004mobile}, or the dynamic nature of the physical world~\cite{fiorini1998motion, liang2021vo}.

One major challenge of global navigation is global localization~\cite{10.1007/978-3-540-24586-5_51}, as the robot needs to constantly know its precise location in the physical world relative to the global map information or coordinate systems. Maintaining this awareness is difficult without accurate GPS sensors or correspondence between the robot's local sensors and the global map. 
While the position estimation can converge as a robot navigates in the environment using continuous noisy GPS or prior location~\cite{sarlin2023orienternet} estimated using Monte Carlo localization~\cite{dellaert1999monte, Miller2021AnyWY}, the convergence could still take a long time if the pose estimation at each time step is unreliable, which can be the bottleneck of the overall performance. 




\begin{figure}[t]
    \centering
    \includegraphics[width=\linewidth]{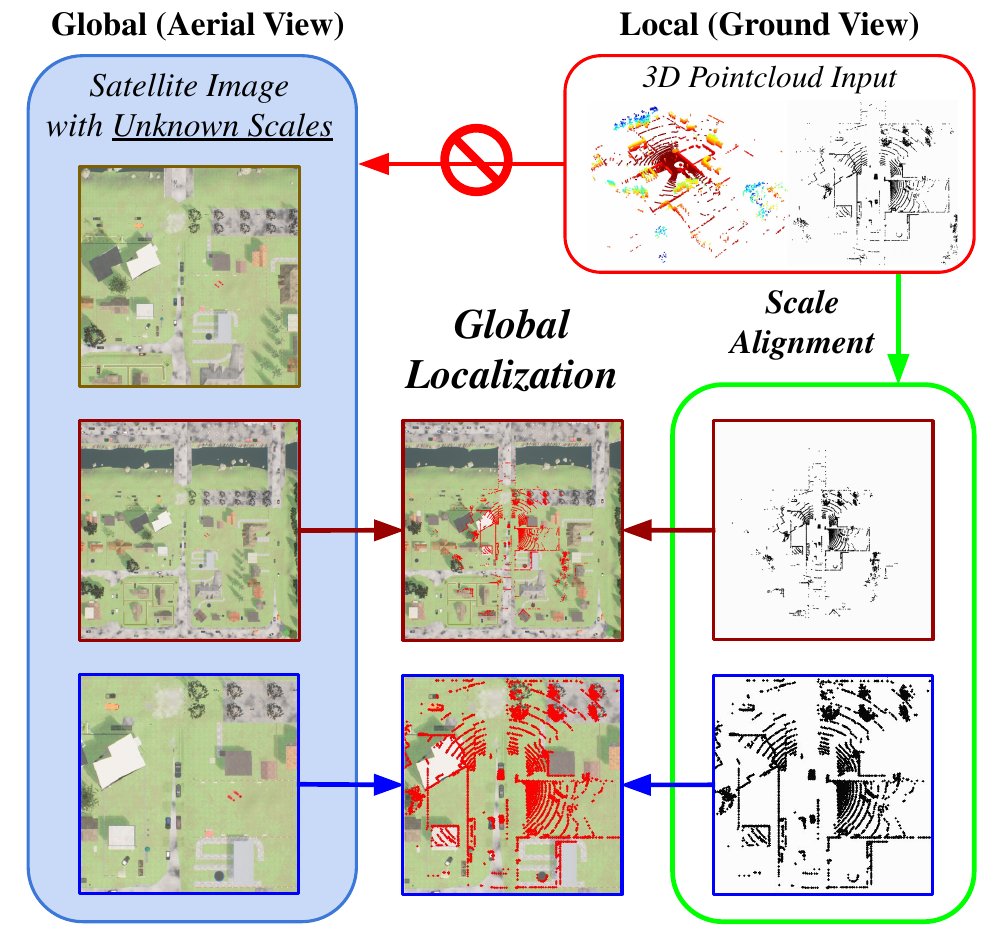}
    \caption{\textbf{Overview of global localization and proposed \ours:} Utilizing a local ground LiDAR and an aerial-view map, our goal is to identify the corresponding position and orientation of ground observations relative to the map. This task presents two significant challenges: cross-modality matching and varying scales of the map. To address these, our method employs a unified network designed to process both point and image modalities, while explicitly managing the scale discrepancies between ground and aerial views. Our assumption of an unknown scale not only distinguishes our method from previous approaches~\cite{Tang2021GetTT, Tang2020RSLNetLI}, but also introduces a more challenging task.
    }
    \label{fig:cover}
    \vspace{-13pt}
\end{figure}

\begin{figure*}[t]
    \centering
    \includegraphics[width=\textwidth]{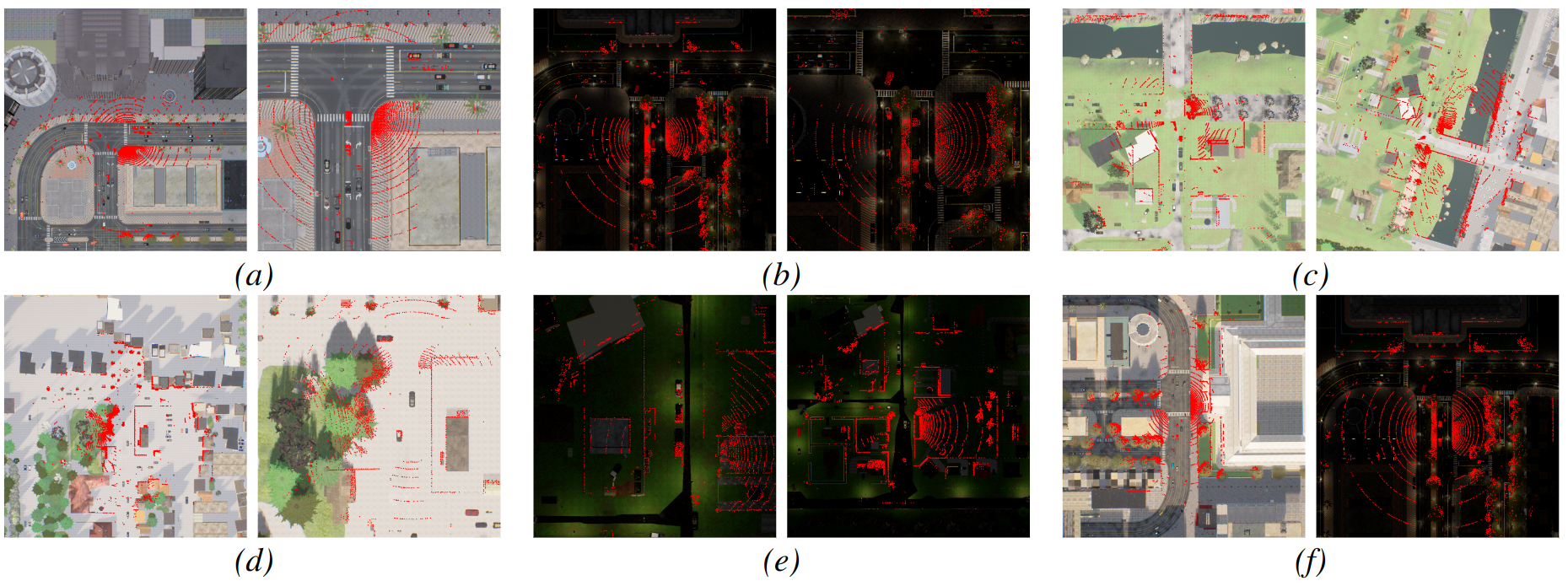}
    \caption{\textbf{Data diversity from CARLA simulator for global localization:} We show the overhead image and LiDAR points in red in their corresponding location. In each pair, we show images on the same area with different scales (a, b, d, e), orientations(c, f), and lighting conditions (f). Since the data from the ground and air might be collected at different times, part of the LiDAR points would not correspond to dynamic objects (cars, etc.) in the aerial view, but static objects (buildings, etc.) can match well with the ground scan.}
    \label{fig:data}
    \vspace{-15pt}
\end{figure*}

An alternative to GPS-based localization involves using local observations to identify a robot's position within a city-scale map. However, this approach introduces several challenges:
First, matching and registration between a local scan and city-wide map is difficult and time-consuming as the overlapping region of those data is small~\cite{Lian2017ACO}. As more map regions are included, the possibility of false positive pose estimations also increases due to increased potential similar match pairs. Second, there is a representation and modality gap between local observation and the global map, and the global map could be out-of-date~\cite{guan2023crossloc3d}, which would not provide consistent and up-to-date local details for accurate matching with real-time local observations in navigation application. Finally, there is a lack of reliable sources to provide accurate global localization ground truth for matching tasks due to the unreliability of GPS data~\cite{gps_manual}.

Our work aims to address these challenges by developing a method that precisely computes the position and orientation of mobile robots or vehicles on the 2D map frame using a single LiDAR observation. While previous studies have explored similar objectives, but often rely on additional information such as accurate semantic segmentation ground truths~\cite{Miller2021AnyWY, 9164147}, access to OpenStreetMap~\cite{sarlin2023orienternet}, or pre-identified landmarks~\cite{8917305} within well-explored areas. Additionally, some methods~\cite{Tang2021GetTT, Tang2020RSLNetLI} process LiDAR data into image patches that match the scale and resolution of overhead maps, aiming to enhance localization accuracy through these adapted representations. Our goal is to overcome these limitations by developing an approach that does not rely on such ideal or specific conditions, making it adaptable for use in a wide range of real-world applications.




\noindent\textbf{Main Contribution:} 
In this paper, we present \ours{} a novel learning-based method for metric-based global localization that leverages LiDAR scans and satellite imagery. As shown in Figure~\ref{fig:cover}, \ours{} tackles the critical challenge of cross-modal matching between these multi-modal sensor data, and unknown scales of the global and local observations. The key contributions of our work include:
\begin{enumerate}
    \item We propose a novel-designed two-stage matching network architecture, \ours{}, to bridge the inherent semantic gap and rectify scale discrepancies between LiDAR and satellite map data. We propose to further match the skeleton feature which offers a more compact and robust representation compared to raw neural features. A novel scale alignment approach is embedded within our method to rectify scale variations between the LiDAR and map data, leading to more accurate and robust localization.
    \item We introduce a scale and skeleton loss function during network training. This loss function plays a crucial role in guiding the network towards learning scale-invariant feature representations. This eliminates the need for pre-processing satellite maps, significantly enhancing real-world applicability in scenarios with unknown map scales and resolutions. 
    \item We create a dataset collected in CARLA~\cite{carla} simulator specifically tailored for global localization utilizing local LiDAR scan and 2D overhead map. Unlike real-world datasets with limited ground truth information, ours leverages simulation data to provide highly accurate ground truth transformations between ground and aerial frames.
    \item We demonstrate \ours{}'s superior performance in camera pose estimation compared to existing state-of-the-art methods~\cite{sarlin2023orienternet} on the KITTI and CARLA datasets. Notably, \ours{} achieves a significant 9.99 meters reduction in average position error on the KITTI benchmark. Furthermore, on the CARLA dataset, it exhibits impressive reductions of 3.79 meters and 25.46\degree{} in position and orientation errors, respectively.
\end{enumerate}

\section{Related Work}

\subsection{Place Recognition}
Achieving global localization in robot navigation necessitates establishing an accurate correspondence between the robot's local sensor measurements and a global map. Place recognition emerges as a prominent technique to address this challenge. It advocates for the creation of a detailed database capturing the environment's visual characteristics before navigation. During navigation, the robot's localization task transforms into a streamlined retrieval process. Extensive research~\cite{guan2023crossloc3d, ma2022overlaptransformer,downes2022city, wu2023pix2map} efficiently searches the database for the scene exhibiting the greatest visual similarity and retrieves the corresponding pose, encompassing both position and orientation within the environment. CrossLoc3D~\cite{guan2023crossloc3d} bridges the representation gap in cross-source 3D place recognition using multi-grained features, adaptive kernels, and iterative refinement for unified metric learning. Pix2Map~\cite{wu2023pix2map} addresses the need for continuous map updates by retrieving the most topologically similar map graph from a database, based on the visual embedding of a given set of test-time egocentric images. 
Our method does not need a pre-defined database and directly estimates the pose based on local data and global map.

\subsection{Metric localization} 


Unlike place recognition, metric localization methods output a fine location estimate or poses of the local data corresponding to global knowledge~\cite{sarlin2023orienternet,wang2023satellite,wang2024fine,sarlin2024snap,camiletto2023u, 9982195}. OrienterNet~\cite{sarlin2023orienternet} achieves sub-meter visual localization through a deep neural network that estimates the location and orientation of a query image by matching a neural Bird’s-Eye View with 2D semantic maps. \cite{wang2024fine} leverages a differentiable spherical transform and a robust correlation-aware homography estimator to achieve sub-pixel resolution and meter-level GPS accuracy by aligning a warped ground image with a corresponding satellite image. 
SNAP~\cite{sarlin2024snap} enhances BEV generation by leveraging ground-level imagery, combining multi-view geometry principles with strong monocular cues. 
While existing research mainly focuses on cross-view geo-localization using RGB images, often making prior assumptions such as perfect semantic segmentation or aligned scales across different views, our work explores cross-modality localization without relying on these assumptions. 


\section{Problem Formulation}
Let query $\mathcal{L}_g$ be a set of 3D points in a single frame captured by a LiDAR sensor $L$ from the ground represented in its relative coordinate in meters. 
Let $\mathcal{M}$ be an overhead map image captured by an RGB camera from the aerial viewpoint, and we take a patch $\mathcal{I}_{map}$ of size $d$ as input based on a prior location provided by a noisy GPS or a prior estimation. Generally, the map coverage $d$ significantly exceeds the GPS error or the vehicle's movement range within a short period.

\noindent\textbf{Task Definition.} \textit{Given a ground scan $\mathcal{L}_g$ in 3D egocentric frame and an RGB map patch $\mathcal{I}_{map}$ in uv-coordiante frame, we want to estimate a location $(u, v)\in\mathcal{R}^2$ and a heading angle $\theta\in (-\pi, \pi]$ with respect to the uv-map frame.}



\noindent\textbf{Unique Setting.} Previous research in place recognition often relies on data from a single platformm~\cite{oxford, kitti} or modality~\cite{guan2023crossloc3d}. In contrast, our approach incorporates a more complex scenario by fusing diverse data, including global aerial camera views and detailed ground scans.  Furthermore, we move beyond the coarse localization achieved by traditional place recognition methods.  Our objective is to achieve fine-grained prediction of both location and orientation, drawing upon both local and global observations.  Existing metric-based localization methods typically address settings with a single modality~\cite{sarlin2023orienternet, sarlin2024snap, guan2023crossloc3d} or require significant pre-processing steps, such as manual scale alignment~\cite{Tang2021GetTT, Tang2020RSLNetLI} or readily available semantic information~\cite{Miller2021AnyWY, 9164147}.  Our approach offers greater flexibility and generalizability by handling raw sensor inputs with varying scales and incorporating real-world factors like noise and temporal discrepancies between observations. 



\section{Simulation Dataset}

One of the major challenges in using real-world data for metric localization is noisy ground truth location. In real-world scenarios, localization with respect to the global frame can only be determined by noisy GPS sensors. To provide more accurate data for training, we use the CARLA~\cite{carla} simulator to collect observations and ground truth locations with respect to the global map frame, as shown in Figure~\ref{fig:data}. Through accurate simulation data, we hope to improve metric localization in the real world.

\noindent\textbf{Data Collection.} We use CARLA 0.9.10 for data collection, which consists of 8 different towns. We use 5 towns for training, and 4 towns for validation and testing, with 1 town overlapping with the training set. The routes in local areas with junctions have an average length of 100 meters, and the routes along curved highways have an average length of 400 meters. In each route and scenario, there are other moving vehicles that interact with the environment.

The ground LiDAR is mounted 1.3 meters in front of the vehicle's center and 2.5 meters above the ground. To simulate satellite images and make sure the image is within the boundary, the RGB camera is mounted 1200 meters above the vehicle. The horizontal and vertical field of view are $20$\degree, resulting in an image with $5000\times5000$ resolution.

We also record the privileged information within the simulation, including the location and orientation of the ego vehicle with respect to the CARLA world frame, and the corresponding transformation matrix. 
We down-sample the local and global observations separately and randomly pick a local-ground pair that belongs to the same town. After down-sampling, we construct 35081 pairs for training and 1385 pairs for validation and testing.




\noindent\textbf{Data Diversity and Augmentation.} To improve data diversity, we consider various factors including map orientation, time lag, scaling, and time of the day.
\begin{itemize}
    \item \textit{Time lag:} The overhead map and ground scan are chosen independently, so the overhead map is not always up to date, just like the satellite map in the real world. We use the ground truth transformation with respect to the world frame to find the ground truth correspondence. 
    \item \textit{Map orientation:} Since the camera for the overhead view is mounted on the vehicle instead of the world frame, the orientation of the map does not guarantee a north-up convention. Since the choice of the overhead map is independent of the ground scan, the map orientation does not affect the ground location or orientation.
    \item \textit{Scaling and resolution:} To increase the diversity of the overhead view scales and resolution, we first determine a scaling factor $s$ and choose a patch of size $s\times d$ before resizing it to size $d$. 
    \item \textit{Transformation and rotation offset:} During training, the center of the overhead image patch is randomly adjusted with translation and rotation so it is away from the ground truth ground location. 
    \item \textit{Time and lighting:} At each time, we randomly adjust the time of the day in the simulation so the overhead images are captured with different lighting conditions.
\end{itemize}

\section{Method}

\begin{figure*}[t]
    \centering
    \includegraphics[width=0.9\textwidth]{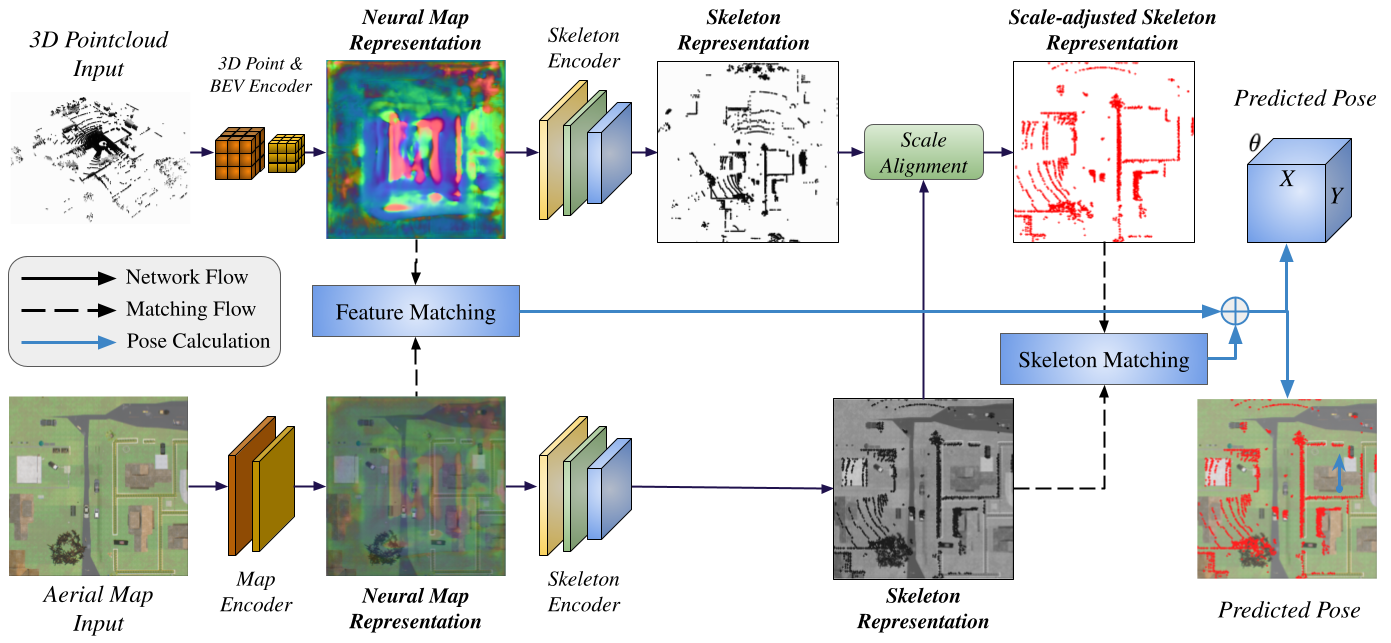}
    \caption{\textbf{Architecture of our proposed network \ours:} \ours{}  processes LiDAR point clouds and aerial maps through separate encoders to generate neural feature representations. These features then undergo a two-stage matching process: initial matching for general correspondence and skeleton-based matching with a predicted scale adjustment to account for potential scale discrepancies. Finally, \ours{}  fuses the results from both stages to generate a robust final estimation score for accurate camera pose determination.}
    \label{fig:network}
    \vspace{-15pt}
\end{figure*}

In this section, we present the details of our proposed \ours{} . It tackles pose estimation by leveraging both LiDAR and RGB map data. As shown in Figure~\ref{fig:network}, we have the following components:
\begin{enumerate*}[label=(\Alph*)]
    \item 3D point cloud encoder.
    \item Map encoder.
    \item Scale Alignment module.
    \item Template matching module for pose estimation.
    \item Training losses.
\end{enumerate*}
Next, we will dive into the details of each module.

\subsection{3D Point Cloud Encoding}
Given a ground LiDAR scan $\mathcal{L}_g$, we want to produce a local BEV feature $f_{bev} \in \mathbb{R}^{H\times W \times C}$ with consistent spatial resolution since the LiDAR scans are always in meter.

\noindent\textbf{Voxelization.} The point cloud is first discretized into  equally spaced 3D voxels $V \in \mathbb{R}^{H \times W \times L}$. Each voxel $v\in\mathbb{R}^{N\times 3}$ encapsulates the points corresponding to its location, where $N$ is the number of points in the voxel. The points are either downsampled or zero-padded to meet the required size.

\noindent\textbf{BEV Features.} Within each voxel, we use a PointNet~\cite{qi2017pointnet} to extract the point features. After that, we used Transformer Blocks~\cite{vaswani2017attention, Guan_2022_WACV} to calculate the mutual relations between the voxels. Similar to PointPillar~\cite{lang2019pointpillars}, the features are scattered back to the original locations and compressed along the height dimension with several 2D convolution layers to obtain a 2D pseudo-image $f_{bev}$ with feature dimension $C$.

\noindent\textbf{Skeleton Mask.} To selectively extract binary skeleton features $f^s_{bev}\in \mathbb{R}^{H\times W \times 2}$ from the BEV feature representation $f_{bev}$, a convolutional neural network (CNN) $\Phi_{bev}^s$ incorporating 2D deformable convolution layers is employed. $\Phi_{bev}^s$ dynamically learns to adapt the receptive field, enabling the network to focus on pertinent regions within $f_{bev}$ that potentially encode crucial skeletal information. We use a skeleton loss to guide this process. 


\subsection{Map Encoding}

\noindent\textbf{Encoding.} Given an RGB map patch $\mathcal{I}_{map}$ from an overhead map image $M$ captured by an RGB camera from the aerial viewpoint, we utilize a pre-trained ResNet-50~\cite{He_2016_CVPR} followed by a VGG19~\cite{Simonyan15} network as a feature extractor to generate a global map feature $f_{map}\in\mathbb{R}^{H\times W \times C}$. 

\noindent\textbf{Map Skeleton.} A convolutional neural network (CNN) $\Phi_{map}^s$ is designed to extract skeletal information $f^s_{map}\in \mathbb{R}^{H\times W \times 2}$ from the processed map features. This network shares a similar design with $\Phi_{bev}^s$ but operates on the map features. It prioritizes specific areas of the map features containing relevant information about the map edges and lines.

\subsection{Scale Alignment}
Unlike the scan encoder where the input is always in the scale of meters, the resolution and scale of the map are uncertain. Addressing the inherent discrepancy in spatial resolution between the local LiDAR scan and the global overhead map necessitates a scale alignment step.

\noindent\textbf{Scale Classifier.} We construct a CNN-based scale classifier $\Phi_{scale}$, which operates solely on the skeletal features $f_{map}^s$ extracted from the map instead of $f_{map}$. This is to avoid the potential issues arising from the redundant geometric information within the neural maps. $\Phi_{scale}$ predicts weights associated with a set of predefined discrete scale bins given a scale range and utilizes a softmax activation function for normalization. The final scale factor $\mathbb{S}$ is then determined by summing the weighted contributions from each bin to obtain a continuous value.

\noindent\textbf{Feature Interpolation.}
The predicted scale factor $\mathbb{S}$ directly reflects the scaling difference between $f_{bev}^s$ and $f_{map^s}$, guided by the scale loss mentioned in Section.~\ref{sec:loss}. Consequently, $f_{bev}^s$ is interpolated using nearest neighbor interpolation based on the estimated scale $\mathbb{S}$. The scaled feature is denoted as $f_{bev}^\mathbb{S}$. 

Specifically, if the predicted scale factor $\mathbb{S}$ is greater than 1, the BEV feature should be reduced in size to match a smaller region of the overhead map. The spatial dimension of the feature $f_{bev}^s$ is reduced proportionally by interpolation and the scaled feature $f_{bev}^\mathbb{S}$ is written on an empty vector with original size of $f_{bev}^s$.
The areas surrounding the scaled feature are filled with zeros. This maintains a consistent spatial size for subsequent processing layers, preventing issues with varying input dimensions. 
On the other hand, $\mathbb{S}$ is less than 1, we focus on a smaller area around the center of $f_{bev}^s$ and enlarge it to the original size.
 This interpolation process establishes consistent spatial resolution between the local LiDAR scan and the global overhead map for effective template matching.

\noindent\subsection{Template Matching}
This subsection details the estimation of a single camera pose $\eta^* = \left( u^*, v^*, \theta^*\right)$ through a two-step matching process: feature matching and skeleton matching. These matching scores are then combined to form a probability distribution that ultimately yields the predicted pose.

\noindent\textbf{Feature Matching.} An exhaustive matching process is conducted between the neural map $f_{map}$ and the BEV feature $f_{bev}$ across various possible camera poses $\eta$. This matching results in a score volume $\Omega$, where each element represents the correlation between $f_{map}$ and a transformed version of $f_{bev}$ according to a specific pose. The transformation, denoted by $\eta(p)$, maps a 2D point $p$ from the BEV space to the corresponding map coordinate frame. Notably, an efficient implementation is achieved by rotating $f_{bev}$ only $N_r$ times and performing a single 2D convolution using batched multiplication in the Fourier domain, as shown in the following equation:
\[
\Omega(\eta) = \frac{1}{XY}\sum_{p\in X \times Y} f_{map}(\eta(p))^T f_{bev}(p),
\]

\noindent\textbf{Skeleton Matching.} Our method extends beyond solely comparing the full neural maps. It incorporates an additional matching step that specifically focuses on the skeletal features extracted from both the BEV representation $f_{bev}^s$ and the map $f_{map}^s$. Similar to feature matching, a matching term $\Psi$ is calculated by exhaustively matching these skeletal features:
\[
\Psi(\eta) = \frac{1}{XY}\sum_{p\in X \times Y} f_{map}^s(\eta(p))^T f^{\mathbb{S}}_{bev}(p),
\]

\noindent\textbf{Probability Scores.} To account for the inherent uncertainty in pose estimation, a discretized probability distribution over camera poses $\eta$ is employed. This involves sampling $N_r$ rotations and constructing a probability volume $P$ considering both the map and BEV feature matching terms $\Omega, \Psi$. The following equation demonstrates how the probability volume is computed using a softmax function:

\begin{equation}
    P = softmax(\Omega + \Psi),
\end{equation}

\noindent\textbf{Pose Estimation.} Finally, the most likely camera pose $\eta$ is determined through maximum likelihood estimation:
\[
\eta^* = argmax_{\eta} P(\eta | f_{bev}, f_{map}, f^{\mathbb{S}}_{bev}, f_{map}^s)
\]

\subsection{Training Loss.}
\label{sec:loss}
Our approach incorporates a novel composite loss function $\mathbb{L}$ to optimize the training process:
\[
\mathbb{L}=L_{uv\theta} + L_{\mathbb{S}} + L_{skeleton}
\]

\begin{figure*}[t]
    \centering
    \includegraphics[width=\textwidth]{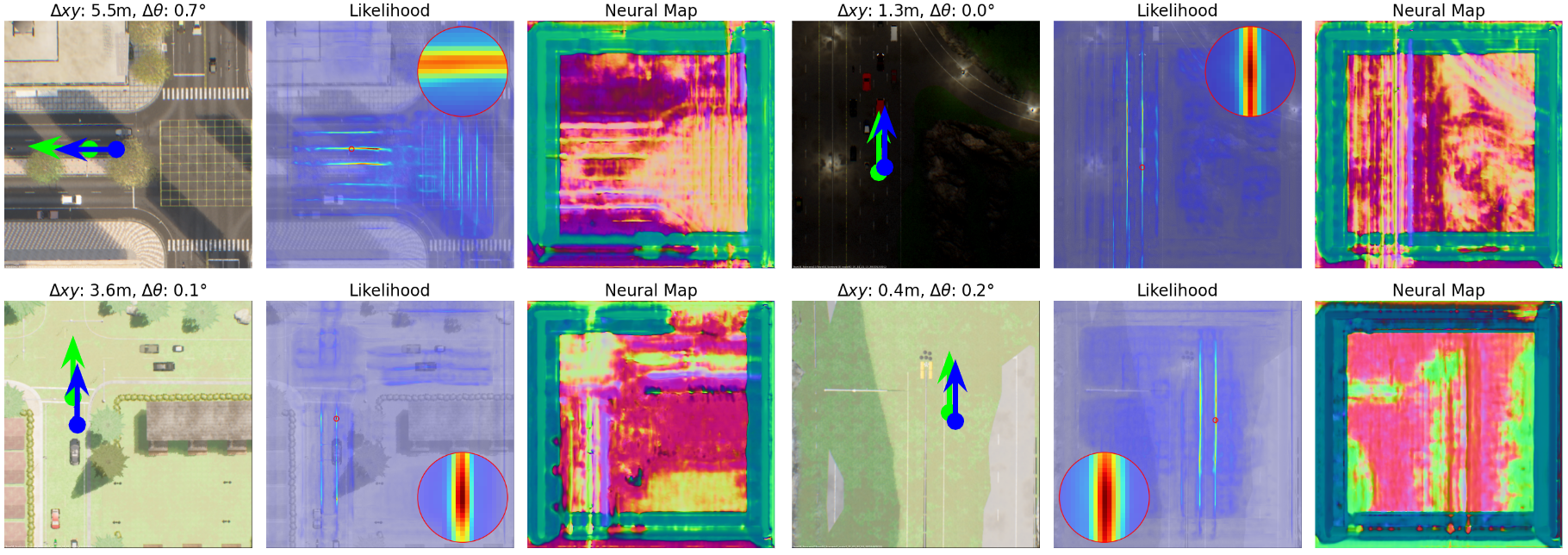}
    \caption{\textbf{\ours{} output visualization in CARLA simulation:} We use green arrow for the ground truth and blue arrow for the predicted pose. We highlight and enlarge the likelihood region near the ground truth in the read circle. Even in case of a larger location error (top left), the pose likelihood distribution have higher value along the lane of the road.}
    \label{fig:result}
    \vspace{-15pt}
\end{figure*}

\noindent \textbf{Pose Loss.} $L_{uv\theta}$ measures the discrepancy between the estimated camera pose $\eta^*$ and the ground truth pose $\eta_{gt}$. We use L2 norm to measure the location error and L1 norm to measure the angle error. Then we employ a discretized probability distribution over $P(\eta)$ potential camera poses and then estimate the pose $\eta^*$ through maximum likelihood. Therefore,  the $L_{uv\theta}$ is formulated as the negative log-likelihood of the predicted pose:
\[
L_{uv\theta}=-log P(\eta^*)
\]
\noindent \textbf{Scale Loss.} \ours{}  employs a dedicated scale classifier to capture the inherent discrepancy in spatial resolution between the LiDAR scan and the overhead map.  This information is crucial for accurate pose estimation. However, directly training the scale classifier solely based on the skeleton features $f_{bev}^s$ and $f_{map}^s$ might not be sufficient, particularly in scenarios where the scale variations are significant. To address this challenge and enhance the model's ability to learn the scale information effectively, we introduce the scale loss $ L_{\mathbb{S}}$. This loss term specifically targets the predicted scale factor $\mathbb{S}$ and the ground truth map scale $\mathbb{S}_{gT}$. $ L_{\mathbb{S}}$ functions as a Mean Squared Error (MSE) loss:
\[
L_{\mathbb{S}} = \frac{1}{B}\sum_i^B (\mathbb{S} - \mathbb{S}_{gt})
\]
where $B$ is the batch size. By minimizing $L_{\mathbb{S}}$, the model is encouraged to predict a scale factor $\mathbb{S}$ that closely aligns with the actual ground truth map scale  $\mathbb{S}_{gT}$.

\noindent \textbf{Skeleton Loss.} $L_{skeleton}$ measures the discrepancy between the predicted skeleton mask $f_{map}^s$ and the ground truth skeleton mask, which often contains crucial information about the edges and corners within the map. Since the skeleton mask is a binary image containing only 0s and 1s, we utilize a Binary Cross Entropy Loss:
\begin{multline*}
L_{skeleton} = -\frac{1}{X \times Y} \sum_{p \in X \times Y} [y_{gt}(p) \cdot log(f_{map}^s(p)) \\
+ (1 - y_{gt}(p)) \cdot log(1 - f_{map}^s(p))]
\end{multline*}
where $p$ is a single 2D point in $f_{map}^s$ and $y$ is the ground truth skeleton mask. Note that the ground truth skeleton mask is obtained by visualizing the LiDAR points on the map, as shown in Figure ~\ref{fig:data}. We predict the skeleton of the entire map patch during inference, and the loss is only calculated on the region with the LiDAR coverage.

\section{Experiments and Evaluations}

\subsection{Dataset and Evaluation Metric}
In addition to data collected in the CARLA~\cite{carla} simulation, we also provide some evaluations on the KITTI~\cite{kitti} dataset for comparison. 
The KITTI dataset is a popular dataset for autonomous driving, collected from a vehicle with LiDAR and camera sensors driving on urban roads. 
The location information, accurate up to within 5 cm, is collected by the OXTS RT3003 GPS sensor. However, according to~\cite{gps_manual}, there are several conditions that need to be satisfied to maintain this level of accuracy, including 15 minutes of continuous operations, an open-sky environment (stay clear from trees, bridges, buildings, or other obstructions) for at least 5 minutes, smooth motion behavior, and so on.

\begin{table*}[t]
  \begin{center}
  \resizebox{\textwidth}{!}{
  \setlength{\tabcolsep}{4pt} 
  \begin{tabular}{ccccccccccccccc}
    \toprule
     \midrule
    \multirow{2}[2]{*}{
    \begin{tabular}{c}
        \textbf{Map} 
    \end{tabular}
    } &   
    \multirow{2}[2]{*}{
    \begin{tabular}{c}
        \textbf{Ground} \\\textbf{Modality}
    \end{tabular}
    } &     
    \multirow{2}[2]{*}{
    \begin{tabular}{c}
        \textbf{Method}\\
    \end{tabular}
    } &     
    \multirow{2}[2]{*}{
    \begin{tabular}{c}
        \textbf{Map Size} \\ (in Pixel)
    \end{tabular}
    } & 
    \multirow{2}[2]{*}{
    \begin{tabular}{c}
        {\textbf{Avg. Loc. / Ori.}} \\ {\textbf{Error} (m / \degree) $\downarrow$}
    \end{tabular}
    } & \multicolumn{3}{c}{\textbf{Lat. R@Xm} $\uparrow$} &  \multicolumn{3}{c}{\textbf{Long. R@Xm} $\uparrow$} &  \multicolumn{3}{c}{{ \textbf{Ori. R@X\degree $\uparrow$}
    }}  \\ \cmidrule(lr){6-8} \cmidrule(lr){9-11}  \cmidrule(lr){12-14}              
 & &  & &  & 1m & 3m 
    & 5m  & 1m & 3m & 5m & 1\degree & 3\degree & 5\degree \\
     \midrule
    \multirow{5}{*}{    
    \begin{tabular}{c}
       OSM\\ (KITTI)
    \end{tabular}
    }  & Image & retrieval &  $256\times 256$ & - / - &  37.47 & 66.24 & 72.89 & 5.94 & 16.88 & 26.97 & 2.97 & 12.32 & 23.27\\
      & Image & refinement &  $256\times 256$ & - / - & 50.83 & 78.10 & 82.22 & 17.75 & 40.32 & 52.40 & 31.03 & 66.76 & 76.07\\
      & Image & OrienterNet &  $256\times 256$ & - / - &  51.26 & 84.77 & 91.81 & 22.39 & 46.79 & 57.81 & 20.41 & 52.24 & 73.53\\
      & LiDAR & OrienterNet* & $256\times 256$ & 28.01 / \textbf{8.30} & 2.86 & 8.80 & 15.37 & 4.03 & 11.47 & 18.48 & 8.62 & 24.11 & 36.38\\
      & LiDAR & \textbf{\ours{}} & $256\times 256$ & \textbf{18.02} / 8.59 & 6.55 & 18.88 & 31.17 & 5.16 & 15.0 & 24.54 & 6.25 & 17.94 & 31.87\\
     \midrule
     \midrule
    \multirow{4}[1]{*}{
        \begin{tabular}{c}
        Satellite\\ (CARLA)
    \end{tabular}
    } & LiDAR & OrienterNet* & $256\times 256$ & 2.23 / 19.15 & 66.87 & 87.67 & 98.61 & 44.38 & 81.51 & 97.69 & 55.78 & 62.71 & 66.26\\
      & LiDAR & \textbf{\ours{}} & $256\times 256$ & \textbf{1.83} / \textbf{3.76} & 76.58 & 92.14 & 100.0 & 44.07 & 88.44 & 99.54 & 57.47 & 68.41 & 74.42\\
       \cmidrule{2-14}
       & LiDAR & OrienterNet* & $1024\times 1024$ & 18.75 / 82.71 & 33.44 & 40.99 & 56.55 & 13.41 & 18.64 & 22.5 & 30.35 & 31.43 & 32.51\\
       & LiDAR & \textbf{\ours{}} & $1024\times 1024$ & \textbf{14.96} / \textbf{57.25} & 50.54 & 57.94 & 69.49 & 18.64 & 24.19 & 29.89 & 42.06 & 44.22 & 46.22\\
     \midrule
  \bottomrule
\end{tabular}
}
\end{center}
\caption{\textbf{Results on KITTI and CARLA:} \ours{} demonstrates significant performance improvements over previous state-of-the-art methods on CARLA datasets. * indicates that we make small modification to the original method to take LiDAR modality input.}
\label{tab:comp1}
\vspace{-3mm}
\end{table*}

\begin{table}[t]
\vspace{-1em}
  \begin{center}
  \resizebox{\linewidth}{!}{
  \setlength{\tabcolsep}{2pt} 
  \begin{tabular}{ccccccccc}
    \toprule
    \multirow{2}[2]{*}{
    \begin{tabular}{c}
        \textbf{Feature} \\\textbf{Match.}
    \end{tabular}
    } &   
    \multirow{2}[2]{*}{
    \begin{tabular}{c}
        \textbf{Skeleton} \\\textbf{Match.}
    \end{tabular}
    } &
    \multirow{2}[2]{*}{
    \begin{tabular}{c}
        \textbf{Scale} \\\textbf{Align.}
    \end{tabular}
    } &
    \multirow{2}[2]{*}{
    \begin{tabular}{c}
        \textbf{Scale} \\\textbf{Aug.}
    \end{tabular}
    } &
    \multirow{2}[2]{*}{
    \begin{tabular}{c}
        \textbf{Avg. Loc. / Ori.}\\ \textbf{Error} (m / \degree) $\downarrow$
    \end{tabular}
    } & \multicolumn{2}{c}{\textbf{Loc. R@Xm} $\uparrow$} &  \multicolumn{2}{c}{{ \textbf{Ori. R@X\degree $\uparrow$}
    }}  \\ \cmidrule(lr){6-7} \cmidrule(lr){8-9}              
        &  &  &  &  & 1m & 3m & 1\degree & 3\degree\\
     \midrule
    \multirow{2}[1]{*}{
    \begin{tabular}{c}
        \checkmark
    \end{tabular}
    } & \multirow{2}[1]{*}{
    \begin{tabular}{c}
        \ding{55}
    \end{tabular}
    }  & \multirow{2}[1]{*}{
    \begin{tabular}{c}
        \ding{55}
    \end{tabular}
    } & \ding{55} & 10.12 / 41.86  & 9.24 & 29.58 & 32.36 & 35.90 \\
    &&& \checkmark & 2.23 / 19.15 & 30.97 & 68.88 & 55.78 & 62.71 \\
    
     \midrule
     
    \multirow{2}[1]{*}{
    \begin{tabular}{c}
        \ding{55}
    \end{tabular}
    } & \multirow{2}[1]{*}{
    \begin{tabular}{c}
        \checkmark
    \end{tabular}
    }  & \multirow{2}[1]{*}{
    \begin{tabular}{c}
        \ding{55}
    \end{tabular}
    } & \ding{55} & 3.53 / 10.61  & 7.55 & 37.29 & 10.32 & 13.56 \\
    &&& \checkmark & 3.40 / 10.26  & 8.47 & 43.76 & 12.17 & 16.18 \\
    
     \midrule
     
    \multirow{2}[1]{*}{
    \begin{tabular}{c}
        \checkmark
    \end{tabular}
    } & \multirow{2}[1]{*}{
    \begin{tabular}{c}
        \checkmark
    \end{tabular}
    }  & \multirow{2}[1]{*}{
    \begin{tabular}{c}
        \ding{55}
    \end{tabular}
    } & \ding{55} & 9.14 / 33.25 & 9.71 & 30.66 & 50.23 & 54.24 \\
    &&& \checkmark & 2.33 / 17.53 & 31.28 & 69.34 & 54.24 & 62.1\\
    
     \midrule
     
    \multirow{2}[1]{*}{
    \begin{tabular}{c}
        \checkmark
    \end{tabular}
    } & \multirow{2}[1]{*}{
    \begin{tabular}{c}
       \checkmark
    \end{tabular}
    }  & \multirow{2}[1]{*}{
    \begin{tabular}{c}
        \checkmark
    \end{tabular}
    } & \ding{55} & 9.50 / 40.41  & 10.94 & 33.59 & 32.67 & 36.52\\
    &&& \checkmark & \textbf{1.83} / \textbf{3.76}  & \textbf{34.05} & \textbf{78.74} & \textbf{57.47} & \textbf{68.41}\\
  \bottomrule
\end{tabular}
}
\end{center}
\caption{\textbf{Ablation studies on different components:}  
``Scaled Feature" refers to utilizing the predicted scale factor to generate the scaled Bird's-Eye View (BEV) feature $f_{bev}^{\mathbb{S}}$. ``Scale augmentation" signifies a data processing step where the overhead map is randomly scaled up or down.}
\label{tab:ablation1}
\vspace{-3mm}
\end{table}

\begin{table}[t]
  \begin{center}
  \resizebox{\linewidth}{!}{
  \setlength{\tabcolsep}{1pt} 
  \begin{tabular}{cccccccc}
    \toprule
    \multirow{2}[2]{*}{
    \begin{tabular}{c}
        \textbf{NLL} \\\textbf{Loss}
    \end{tabular}
    } &   
    \multirow{2}[2]{*}{
    \begin{tabular}{c}
        \textbf{Scale} \\\textbf{Loss}
    \end{tabular}
    } &
    \multirow{2}[2]{*}{
    \begin{tabular}{c}
        \textbf{Skeleton} \\\textbf{Loss}
    \end{tabular}
    } &
    \multirow{2}[2]{*}{
    \begin{tabular}{c}
        \textbf{Avg. Loc. / Ori.}\\ \textbf{Error} (m / \degree) $\downarrow$
    \end{tabular}
    } & \multicolumn{2}{c}{\textbf{Loc. R@Xm} $\uparrow$} &  \multicolumn{2}{c}{{ \textbf{Ori. R@X\degree $\uparrow$}
    }}  \\ \cmidrule(lr){5-6} \cmidrule(lr){7-8}              
        &  &   &  & 1m & 3m & 1\degree & 3\degree \\
     \midrule
     
    \checkmark & \ding{55} & \ding{55} & 2.23 / 19.15 & 30.97 & 68.88 & 55.78 & 62.71\\
    
    \checkmark & \checkmark & \ding{55} & 3.04 / 26.07 & 30.35 & 67.18  & \textbf{59.48} & 64.56 \\
    \checkmark & \ding{55} & \checkmark & 2.67 / 22.38 & 31.28 & 67.80 
 & 59.17 & 64.10\\
    \checkmark & \checkmark & \checkmark & \textbf{1.83} / \textbf{3.76} & \textbf{34.05} & \textbf{78.74}  & 57.47 & \textbf{68.41}\\

  \bottomrule
\end{tabular}
}
\end{center}
\caption{\textbf{Ablation studies on different loss:} Combining both the scale loss ($L_\mathbb{S}$) and skeleton loss ($L_{skeleton}$) leads to superior performance compared to using only one loss function at a time. }
\label{tab:ablation2}
\vspace{-3mm}
\end{table}

\noindent\textbf{Evaluation Metrics.} We used both recall and metric distance for evaluation. \textit{Recall@Xm/\degree} is defined as the percentage of the predicted poses that are within X meter/degree from the true poses. Position recall is calculated at distances of 1, 3, and 5 meters, and orientation recall is calculated at 1, 3, and 5 degrees. For metric distance, we provide the average error between the predicted pose and the ground truth pose in meters and degrees, respectively.

\subsection{Implementation Details}

\noindent\textbf{Preparation.} We first voxelize the points into pillars of size $[2, 2, 30]$, where the range of points in the x, y, and z axes are $[-100, 100], [-100,  100], [-10, 20]$, respectively, and the maximum number of points within a voxel is 128. For the overhead image, we choose patch sizes of $(1024, 1024)$ and $(256, 256)$ with different random scales to obtain different area coverages that also contain the ground truth poses.

\noindent\textbf{\ours{}.} We set the PointNet feature dimension to $16$ and then use $2$ attention blocks with 4 heads and a feature dimension of $16$. After $4$ Conv2D layers with a kernel of size $1$, the matching dimensions of $f_{bev}$ are reduced to $8$.
$\Phi_{bev}^s$ and $\Phi_{map}^s$ have the same structure with $3$ deformable Conv2D with ReLU activation followed by a linear layer. The scale classifier $\Phi_{scale}$ consists of $3$ deformable Conv2D with ReLU activation followed by $2$ linear layers, with 33 scale bins and range of $[0.5, 10]$.

\noindent\textbf{Training Parameters.} Models are trained on $4$ NVIDIA RTX A6000s for $200k$ iterations. We use the Adam optimizer with a learning rate of $0.0001$. We use batch sizes $48$ and $4$ patch sizes $1024$ and $256$, respectively.

\subsection{Results}
This section presents the experimental evaluation and visualizations (Figure~\ref{fig:result}) of our proposed method (\ours{}) on the KITTI and CARLA datasets. We are not able to compare with~\cite{Tang2020RSLNetLI, Tang2021GetTT, doi:10.1177/02783649211045736, 10160267} due to limited access to the source code. We plan to release the code for \ours{}.



\noindent \textbf{Comparisons on CARLA.}
\ours{} demonstrates superior performance in pose estimation compared to OrienterNet~\cite{sarlin2023orienternet}. While OrienterNet originally utilizes ground images as input, we adapted its code to operate on LiDAR data to ensure a fair comparison.

As shown in Table~\ref{tab:comp1}, \ours{}  achieves demonstrably better results on the CARLA dataset. Specifically, it achieves a 0.4-meter reduction in average position error ($u,v$) and a 15.39\degree{} decrease in orientation error when using a  $224 \times 224$ pixel map size. These improvements highlight the effectiveness of \ours{} 's two-stage matching process. Skeleton matching further refines the pose estimation, leading to increased accuracy. Furthermore, \ours{} maintains its advantage as the map size scales up to $1024\times 1024$ pixels. 

\noindent \textbf{Comparisons on KITTI.}
Our evaluation of the KITTI dataset, as detailed in Table~\ref{tab:comp1}, reveals a significant performance difference between metric localization using LiDAR scans and overhead maps compared to methods using image-map pairs. This disparity stems from three primary factors: first, LiDAR data presents a sparse point cloud, especially at longer distances. This sparsity can limit the number of features available for matching with the map, particularly in areas with minimal objects or details. Second,  LiDAR primarily captures 3D geometry, lacking the rich texture and color information offered by ground images. This can hinder matching in scenarios with repetitive structures or similar geometric features. Additionally, KITTI focuses on urban settings with buildings and well-defined structures. While LiDAR excels at capturing 3D geometry, it might be less effective compared to ground images in these environments.

Our proposed method, \ours{} , achieves encouraging results on the KITTI dataset. Compared to OrienterNet, \ours{}  demonstrates a significant improvement of 9.99 average meter in position error. However, it's important to acknowledge a slight increase in orientation error when using KITTI data. This could be attributed to the inherent limitations of the KITTI dataset itself. As mentioned earlier, LiDAR data in KITTI might contain calibration inconsistencies and outliers from GPS, which can affect the performance of our skeleton matching stage. This observation further underscores the importance of using high-quality, well-calibrated datasets like CARLA. With data collected in CARLA, both our proposed \ours{} and Orienter can maintain reasonable performance in both position and orientation estimation tasks. 

\subsection{Ablation Study}
We use map of size $256 \times 256$ for all ablations.

\noindent \textbf{Template Matching.} 
Table~\ref{tab:ablation1} evaluates the contribution of different matching components within \ours{}. The scale augmentation significantly reduces pose estimation error. By exposing the model to maps with different scales during training, it can better learn the visual features and semantic meaning within the map, leading to improved localization accuracy. Compared to feature matching alone, skeleton matching demonstrates greater robustness to scale variations. This is because skeleton information focuses on key edges and corners, which are less affected by scale changes. The proposed scale alignment process, which involves learning a scale factor and scaling the BEV skeleton feature, achieves the best performance. This highlights the critical role of learning scale information. By accurately estimating the scale, the model can perform more effective feature and skeleton matching, ultimately leading to more accurate camera pose estimation.

\noindent \textbf{Training Loss.}
As shown in Table~\ref{tab:ablation2}, our proposed approach that utilizes both the scale loss $L_{\mathbb{S}}$ and skeleton loss $L_{skeleton}$ achieves superior performance compared to using only one loss function at a time. This highlights the importance of jointly considering both aspects during training. The scale loss focuses on constraining the scale of the features extracted from the LiDAR scan to better align with the features extracted from the overhead map. However, solely relying on $L_{\mathbb{S}}$ might be insufficient. The key insight is that the scale of the features is inherently linked to the skeleton information. If the skeleton features are inaccurate (due to the absence of $L_{skeleton}$, the scale learned by $L_{\mathbb{S}}$ might also be inaccurate, leading to suboptimal pose estimation. Therefore, by combining both $L_{\mathbb{S}}$ and $L_{skeleton}$, the model can learn not only to match features at the appropriate scale but also to extract meaningful skeleton information that refines the scale estimation process. 

\section{Conclusions, Limitations, and Future Work}
In this paper,  we present \ours{}, a novel approach for accurate global localization using LiDAR and satellite maps. \ours{} tackles two key challenges: efficiently matching LiDAR data with satellite imagery, and handling scale discrepancies between these two modalities. Our solution lies in a unified network architecture and a unique scale and skeleton loss function. These advancements enable \ours{} to achieve superior pose estimation performance and eliminate the need for map pre-processing, making it adaptable to real-world scenarios.

While our simulation results are promising, future efforts will focus on bridging the gap to real-world applications through domain adaptation techniques. Additionally, we plan to release the code, data, and ground truth alignment methods to foster further research in global localization using LiDAR and satellite maps.

\noindent\textbf{Acknowledgement} 
This work was supported in part by ARO Grants  W911NF2310046, W911NF2310352,  and U.S. Army Cooperative Agreement W911NF2120076.



{\small
\bibliographystyle{IEEEtran}
\bibliography{citation}
}

\end{document}